%% file: arxiv_sigconf.tex
\documentclass[sigconf,natbib=true]{acmart}
\AtBeginDocument{%
  }

\setcopyright{acmlicensed}
\copyrightyear{2018}
\acmYear{2018}
\acmDOI{XXXXXXX.XXXXXXX}
\acmConference[Conference acronym 'XX]{Make sure to enter the correct
  conference title from your rights confirmation email}{June 03--05,
  2018}{Woodstock, NY}
\acmISBN{978-1-4503-XXXX-X/2018/06}

\usepackage{algorithm}
\usepackage{algpseudocode}
\usepackage{multirow}
\usepackage{enumitem}
\usepackage{booktabs}
\usepackage{amsmath}
\usepackage[
        noabbrev,
        capitalise,
        nameinlink,
    ]
    {cleveref}
\usepackage{tabularx} 
\usepackage{longtable}
\usepackage{cancel}
\usepackage{subfigure}
\begin{document}

\title{KGLA: Knowledge Graph Enhanced Language Agents for Recommendation}



\author{Taicheng Guo}
\authornote{Work was done as an intern at Amazon.}
\authornote{Both authors contribute equally to the work.}
\affiliation{
    \institution{University of Notre Dame}
    \country{}
}
\email{tguo2@nd.edu}

\author{Chaochun Liu}
\authornotemark[2]
\affiliation{
    \institution{Amazon}
    \country{}
}
\email{chchliu@amazon.com}

\author{Hai Wang}
\affiliation{
    \institution{Amazon}
    \country{}
}
\email{ihaiwang@amazon.com}

\author{Varun Mannam}
\affiliation{
    \institution{Amazon}
    \country{}
}
\email{mannamvs@amazon.com}

\author{Fang Wang}
\affiliation{
    \institution{Amazon}
    \country{}
}
\email{fwfang@amazon.com}

\author{Xin Chen}
\affiliation{
    \institution{Amazon}
    \country{}
}
\email{xcaa@amazon.com}

\author{Xiangliang Zhang}
\affiliation{
    \institution{University of Notre Dame}
    \country{}
}
\email{xzhang33@nd.edu}

\author{Chandan K. Reddy}
\affiliation{
    \institution{Amazon}
    \country{}
}
\email{ckreddy@amazon.com}

\begin{abstract}
Language agents have recently been used to simulate human behavior and user-item interactions for recommendation systems. However, current work focuses only on simulating interactions but neglects the potential rationales behind why the interaction happens, leading to inaccurate simulated user profiles and ineffective recommendations. In this work, we explore the utility of Knowledge Graphs (KGs), which contain extensive and reliable information about relationships between users and items, for recommendation. Our key insight is that the paths in a KG can represent rationales behind why user interacts with items, eliciting the underlying reasons for user preferences and enriching user profiles. Leveraging this insight, we propose \textit{Knowledge Graph Enhanced Language Agents (KGLA)}, a framework that unifies language agents and KG for recommendation systems. In the simulated recommendation scenario, we position the user and item within the KG and incorporate KG paths as natural language descriptions into the simulation. This allows language agents to interact with each other and uncover sufficient rationale behind their interactions, making the simulation more accurate and aligned with real-world cases, thus improving recommendation performance. Our experimental results show that KGLA significantly improves recommendation performance (with a 33\%-95\% boost in NDCG@1 among three widely used benchmarks) compared to the previous best baseline method.
\end{abstract}



\renewcommand{\shortauthors}{Guo et al.}

\maketitle

\input{1-Introduction}

\input{2-Related_work}
\input{3-Methodology}
\input{4-Experiment}

\input{5-Conclusion}

\input{6-Appendix_short}
\bibliographystyle{ACM-Reference-Format}
\bibliography{sample-base}

\end{document}

%% file: 1-Introduction.tex
\section{Introduction}

\begin{figure*}[t]
\begin{center}
    \subfigure[Without KG]{
        \includegraphics[scale=0.43]{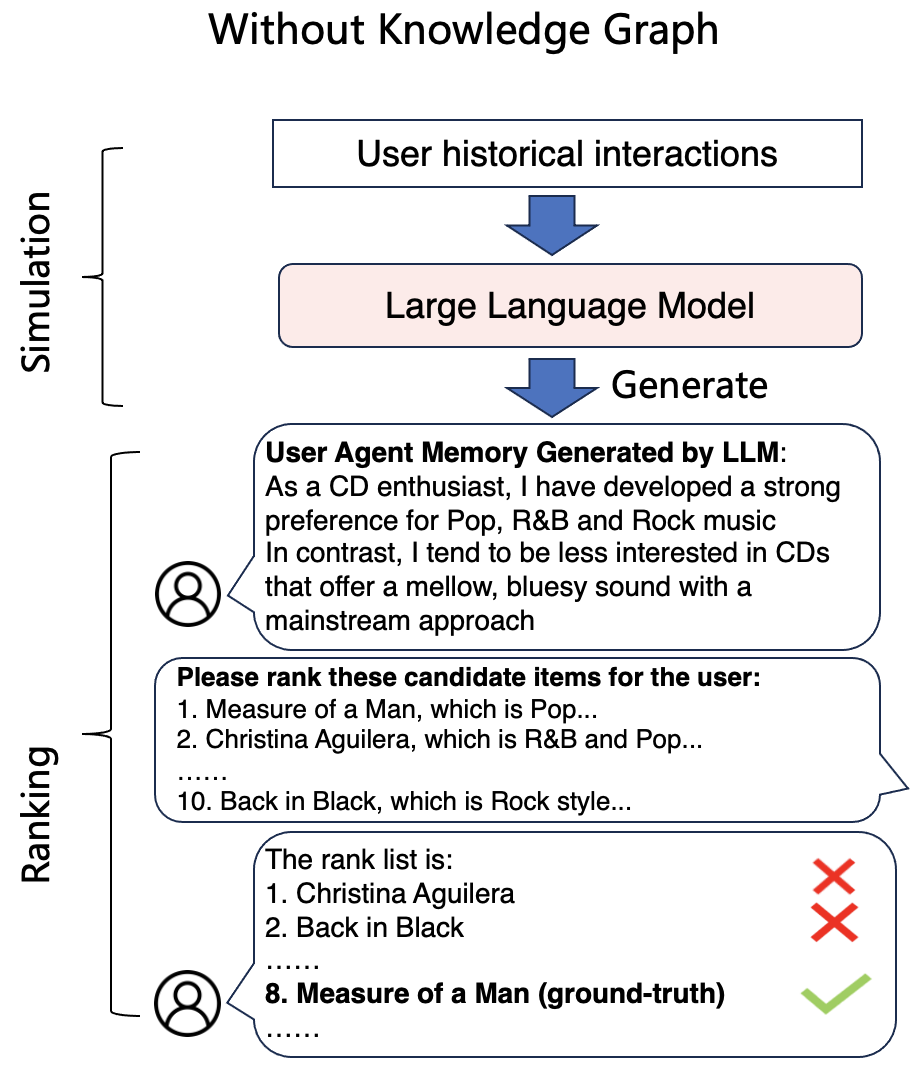} 
        \label{fig:intro_left}
    }
    \hspace{0.5cm} 
    \subfigure[With KG]{
        \includegraphics[scale=0.43]{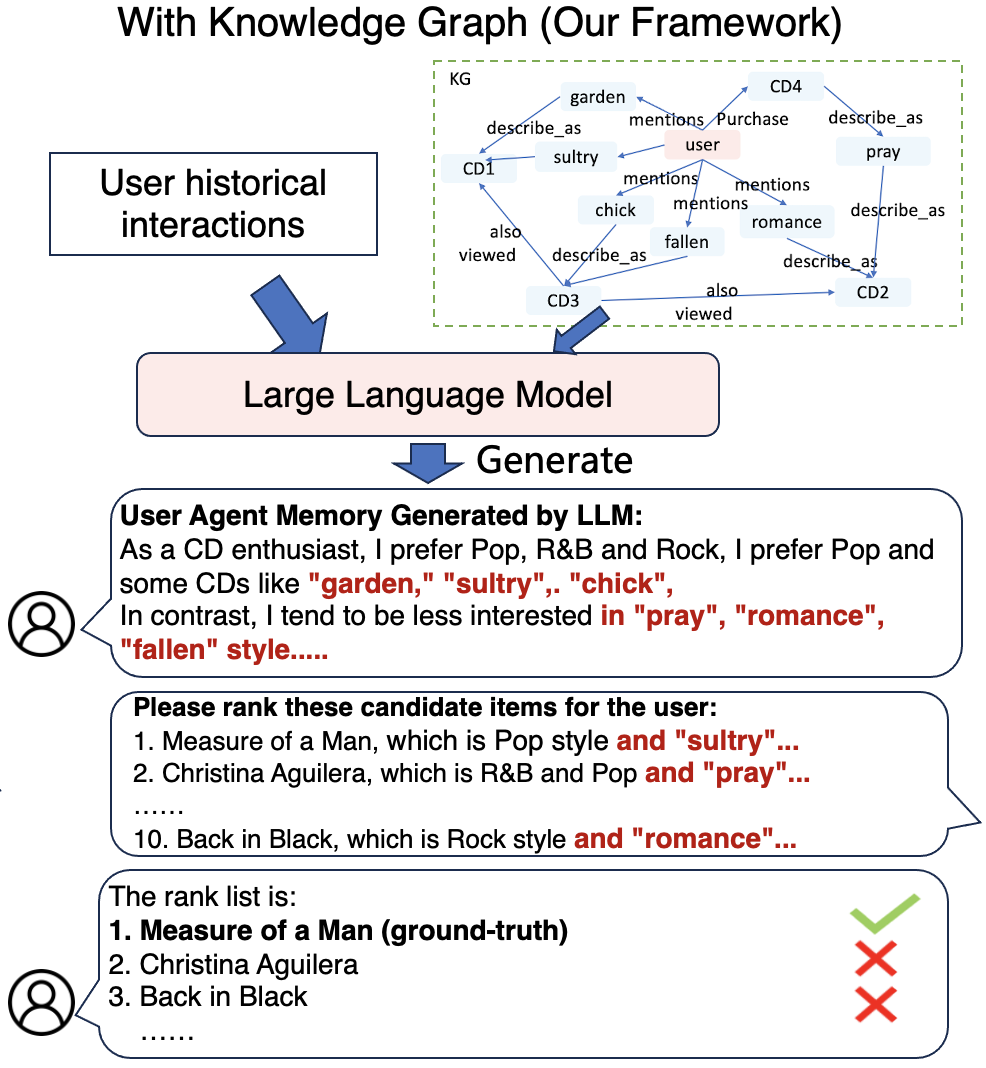} 
        \label{fig:intro_right}
    }
\end{center}
\vspace{-10pt}
\caption{Examples of user agent memory generated (a) without KG and (b) with KG. The user agent memory generated without KG only contains some general descriptions, while the memory generated with KG includes more specific terms (highlighted in red), demonstrating more precise user preferences.}
\label{fig:intro}
\vspace{-10pt}
\end{figure*}

Large Language Model (LLM) agents have demonstrated remarkable capabilities in conversation and decision-making across various tasks~\cite{xi2023rise, guo2024large, kemper2024retrieval, zhang2024lorec, sun2024large, 10.1145/3626772.3657690}. By enabling multiple LLM agents to cooperate, each agent can possess its own profile, actions, and behaviors, simulating diverse human behaviors. Recent works have employed LLM Agents to simulate real-world recommendation scenarios~\cite{zhang2024agentcf, wang2024multi}. In these simulations, each agent represents a user or an item in the recommendation system, maintaining a memory that records the user or item's profiles. The interactions between agents simulate user interactions with items, dynamically updating the agents' memories to reflect changes in user preferences or the recognition of unique item features.

While LLM Agents can capture more explainable and explicit user profiles through text during the recommendation process, previous work has only considered adding brief and simplistic descriptions of users and items as prompts during the interactions. Consequently, the updated memory for user agents is often non-specific and general due to a lack of rationalized information about the user's choices (see  \Cref{fig:intro}). With these inadequate memory profiles, LLMs struggle to identify precise user preferences and may recommend irrelevant items. In this work, we analyze the main cause of insufficient profiles for user and item agents, which primarily arises from simulations relying solely on simple, superficial descriptions without rationalizing why users like or dislike certain items. As a result, most user or item agent memories are generated by LLMs based on limited context, heavily relying on the pre-trained knowledge of LLMs and making them susceptible to generating generic profiles. User profiling is a critical task for recommender systems. Hence, addressing the question of \textit{`how to provide user agents with sufficient information during Agent simulation to obtain more rational and precise user profiles?'} remains a crucial but unresolved problem.

To address this issue, we aim to utilize an existing Knowledge Graph (KG) containing entities with specific meanings to provide detailed rationales to enhance agent simulation for recommendation. For example, paths in a KG can explain why a user may like an item ($\text{User} \stackrel{\text{mentions}}{\longrightarrow} \text{features} \stackrel{\text{describe\_as}}{\longrightarrow} \text{CD}$), thereby helping to build better user and item agent profiles. Most prior work on using KGs for recommendation encodes knowledge from a KG into embeddings and trains a graph neural network to generate graph embeddings, which are then concatenated with user embeddings to represent user profiles. In contrast, since we employ LLM Agents to simulate users and items when the number of interactions is limited and profiles are represented as text, KG information should also be represented in text form to effectively influence the agent simulation process.

Based on this motivation, we first identify the information from the KG required for recommendation. We formulate this as a `\textit{path-to-text}' problem, where users and items are treated as nodes in a KG, and paths between these nodes represent the rationales behind a user’s choice of an item. These paths help build more precise user agent memory. While a few previous works have focused on leveraging KGs for LLM agents~\cite{jiang2024kg,xu2024generate,luo2023reasoning}, to the best of our knowledge, most existing KG+Agent methods focus primarily on the question-answering tasks. In such methods, given a question, the agent starts from a node in the KG and traverses the graph deductively to find the answer node. In contrast, our setting assumes that both the start and destination entities (i.e., the user and item) in the KG are already known. Instead of traversal, the LLM agent needs to analyze paths between the nodes and summarize key rationales. To achieve this, we propose \textit{KGLA: Knowledge Graph Enhanced Autonomous Language Agents}, a framework composed of three modules — Path Extraction, Path Translation, and Path Incorporation — that extract, translate, and incorporate KG paths to provide faithful rationale information. These rationales improve agent profiles and enhance recommendation accuracy. Based on our proposed framework, we conduct an in-depth analysis to demonstrate how it improves agent simulation for recommendation, thereby boosting recommendation performance. To summarize, our \textbf{key contributions} are:
\begin{itemize}[leftmargin=*]

    \item To the best of our knowledge, we are the first to leverage KGs and demonstrate the effectiveness of KGs in enhancing agent profiles for recommendation.
    \item We identify key challenges and propose a novel framework to extract, translate, and incorporate KG information into LLM agent-based recommendations. Our KG-enhanced framework provides interpretable explanations for the recommendation process and is also applicable to a broad range of KG+Agent simulation tasks.
    \item We demonstrate the effectiveness of our method through experiments on various public datasets. The results show that our method consistently outperforms baselines (achieving relative improvements of 95.34\%, 33.24\%, and 40.79\% in NDCG@1 on three widely used benchmarks, compared to the previous best baseline), and the agent memories are more precise than prior approaches. Our method offers substantial practical benefits, especially in recommendation scenarios with limited interactions. By modeling user preferences through text, it complements existing deep learning-based systems, provides explainable recommendations and enables users to independently update their profiles for more interactive recommendations.
\end{itemize}

%% file: 2-Related_work.tex
\section{Related Work} 
\subsection{LLM Agents}

Large language model-based agents are widely researched for various scenarios. These agents can take actions, interact with the environment, and obtain feedback. The key capabilities of LLM agents are: 1) \textit{Take Action}~\cite{yao2022react}: Prompting the agent to perform an action or make a decision; 2) \textit{Reflection}~\cite{shinn2023reflexion}: After taking an action, the agent receives feedback from the environment, which allows it to reflect on the task and improve decision making in subsequent trials; and 3) \textit{Memory}~\cite{zhang2024survey}: Agents store the lessons learned from reflection. In our work, during the simulation stage, the user agent first selects a preferred item from a list of items (Taking Action). Given the ground-truth item, the agent then reflects on its previous choice (Reflection). Subsequently, it updates its profile to record its preference (Memory). In ranking stage, the user agent provides a preferred rank list of items (Taking Action) based on the given list. 

\subsection{LLM Agents for Recommendation}

Due to the strong capabilities of LLM agents, they have successfully been employed in recommendation tasks.~\cite{wang2024multi} proposes two categories in agent-based recommendation: recommender-oriented and simulation-oriented. The recommender-oriented approach aims to develop a recommender agent equipped with enhanced planning, reasoning, memory, and tool-using capabilities to assist in recommending items to users. RecMind~\cite{wang2023recmind} and InteRecAgent~\cite{huang2023recommender} both develop single agents with improved capabilities for recommendation. RAH~\cite{shu2023rah} and MACRec~\cite{wang2024multi} support collaboration among different types of agents to recommend items to users. ~\cite{shi2024enhancing} proposes a learnable LLM planner framework for long-term recommendation.

In contrast, the simulation-oriented approach focuses on using agents to simulate individual user behaviors and item characteristics. RecAgent~\cite{wang2023user} and Agent4Rec~\cite{zhang2024generative} employ agents as user simulators to simulate interactions between users and recommender systems, investigating the plausibility of simulated user behaviors. AgentCF~\cite{zhang2024agentcf} explores simulating user-item interactions using user agents and item agents. The memories of user agents and item agents are dynamically updated during the simulation, and the final memory is used to recommend items to users.

Our work falls under the category of simulation-oriented approaches. Previous simulation work only focused on using agent memory to simulate user profiles for the recommendation, neglecting the construction of high-quality user agent memories. Poor user agent memory can impair recommendation performance. \textit{Our work leverages KGs to incorporate comprehensive rationale information into the entire process of LLM Agents-based recommendation}, thereby building well-grounded and precise user agent memories.

\subsection{Knowledge Graph and LLM Agents}

With the latest advances in LLMs, they are utilized in synergy with KGs to provide accurate information. Most previous works~\cite{jiang2024kg,xu2024generate,luo2023reasoning} focus on synergizing LLM and KGs for question-answering tasks. The synergizing paradigm is deductive - Given a question, LLMs identify the starting entity node and then act as planners to generate relation paths, traversing the KG to reach potential entities that can answer the question. However, the synergizing paradigm between LLMs and KGs in recommendation systems is inductive and differs from previous methods. In this context, the starting entity node (user) and the target entity node (item) are already known within the KG. LLMs are responsible for analyzing all the paths between these two nodes and summarizing the key reasons why the user node can reach the target node. These rationales are then used to update the user agent's memory. In contrast to previous work, \textit{our approach is the first to leverage LLM agents to inductively analyze KG paths for reflection, thereby enhancing recommendation performance.}

%% file: 3-Methodology.tex
\section{Methodology}

\begin{figure*}[tp]
\begin{center}
\includegraphics[scale=0.45]{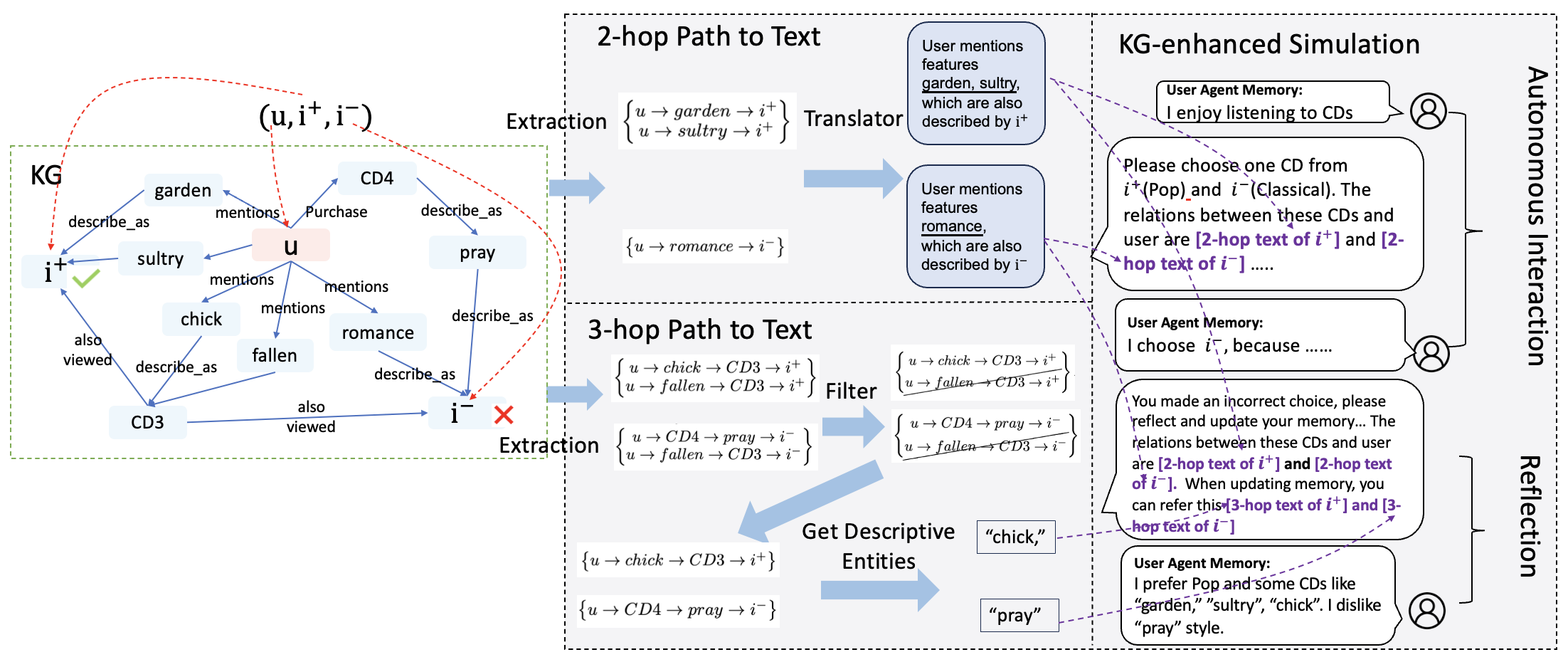}
\end{center}
\caption{The framework of our proposed KG-enhanced Agent Simulation for recommendation. Given $(u, i^+, i^-)$, our framework can retrieve and translate KG information which would be incorporated into the Simulation, guiding LLM agents to analyze the possible reasons for user choices based on KG, summarize the user's precise preferences, and update the user agent's memory.}
\label{fig:framework}
\vspace{-10pt}
\end{figure*}

\subsection{Notations and Problem Definition}
We first introduce the required notations and define the problem.

\paragraph{Recommender System.} Let the user set be denoted by $U$, and the item set be denoted by $I$. Each user $u \in U$ has an interaction history $[i_1, i_2, \ldots, i_n]$. We have a ground-truth label $y$ indicating whether the user likes an item ($y=1$) or does not like it ($y=0$).

\paragraph{Knowledge Graph.} A KG is a structured representation of knowledge that contains abundant facts in the form of triples $G = \{(e_h, r, e_t) \mid e_h, e_t \in \mathcal{E}, r \in \mathcal{R}\}$, where $e_h$ and $e_t$ are head and tail entities, respectively, and $r$ is the relation between them. $\mathcal{E}$ is the entities set in the KG and $\mathcal{R}$ is the relations set in the KG.
A path in a KG is a sequence of triples: $p = e_0 \xrightarrow{r_1} e_1 \xrightarrow{r_2} \ldots \xrightarrow{r_l} e_l$, connecting the entity $e_0$ to the entity $e_l$.

\paragraph{LLM Agent and LLM Recommendation.} For the LLM agent, we denote its memory as $M$. During the simulation stage, the reflection process for the agent is represented by the function $Reflection$. In the ranking stage, given a user and a list of items, the LLM used to generate the ranked list is represented by the function $LLM$.

\subsection{Overall Framework}
\label{sec:frame}
The architecture of our proposed framework (see \Cref{fig:framework} and \Cref{alg:framework})  has three stages: \textbf{Initialization}, \textbf{Simulation} and \textbf{Ranking}. It is designed to combine LLM agents with KGs to enhance agent memory during simulation and improve recommendation performance during ranking. In the \textit{initialization stage}, the memories of all users are set using the template ``I enjoy ...," while item memories are initialized with the titles and categories of the items.

\begin{algorithm}[h]
\caption{KG-enhanced Recommendation}\label{alg:framework}
\begin{algorithmic}[1]
\Require Knowledge Graph $G$, user $u$, training samples for $u$: 
\[
\{(u, i_1^+, i_1^-), \ldots, (u, i_n^+, i_n^-)\}
\]
where $i^+$ is a positive item and $i^-$ is a negative item sampled during autonomous interaction. Testing samples for $u$: 
\[
\{c_1, \ldots, c_n\}
\]
Functions \texttt{Trans-2HOP} and \texttt{Trans-3HOP} which translate 2-hop and 3-hop paths between $u$ and $i$ to text for recommendation.  
\Ensure Recommend rank list $R$ to user $u$. 
\State \textbf{Stage 1: Initialization} 
\State Initialize user memory $M_u$ and item memory $M_i$ for each item.
\State \textbf{Stage 2: Simulation}
\For{each $(i_k^+, i_k^-)$ in training samples}
    \State \parbox[t]{\linewidth}{
    $T_2^{ui^+}, T_3^{ui^+}  \gets \texttt{Trans-2HOP}(G, u, i^+), \texttt{Trans-3HOP} \\(G, u, i^+, \{(u, i_1^+, i_1^-), \ldots, (u, i_n^+, i_n^-)\})$
    }
    \State \parbox[t]{\linewidth}{
    $T_2^{ui^-}, T_3^{ui^-} \gets \texttt{Trans-2HOP}(G, u, i^-), \texttt{Trans-3HOP} \\(G, u, i^-, \{(u, i_1^+, i_1^-), \ldots, (u, i_n^+, i_n^-)\})$
    }
    \State // Autonomous Interaction
    \State $i_{\text{select}}, y_{\text{exp}} \gets f_{LLM}(M_u, M_{i^+}, M_{i^-}, T_2^{ui^+}, T_2^{ui^-})$
    \State // Reflection
    \State \parbox[t]{\linewidth}{
    $M_u \gets \texttt{Reflection}(i_{\text{select}}, y_{\text{exp}}, M_{i^+}, M_{i^-}, \\ T_2^{ui^+}, T_2^{ui^-}, T_3^{ui^+}, T_3^{ui^-})$
    }
    \State \parbox[t]{\linewidth}{
    $M_i^+, M_i^- \gets \texttt{Reflection}(i_{\text{select}}, y_{\text{exp}}, M_{i^+}, \\ M_{i^-}, T_2^{ui^+}, T_2^{ui^-}, T_3^{ui^+}, T_3^{ui^-})$
    }
\EndFor
\State \textbf{Stage 3: Ranking}
\State \parbox[t]{\linewidth}{
    $\{T_2^{c_1}, \ldots, T_2^{c_n}\} \gets \{\texttt{Trans-2HOP}(u, c_1), \ldots, \\  \texttt{Trans-2HOP}(u, c_n)\}$
}
\State $R \gets f_{LLM}(M_u, \{M_{c_1}, \ldots, M_{c_n}\}, \{T_2^{c_1}, \ldots, T_2^{c_n}\})$
\State Recommend rank list $R$ to user $u$.
\end{algorithmic}
\end{algorithm}

The \textit{simulation stage} consists of two phases: Autonomous Interaction and Reflection. Given a user $u$ with a chronological sequence of behavioral interactions $[i_1, i_2, \ldots, i_n]$, the simulation stage aims to optimize the memories representing the user and item profiles by simulating real-world user-item interactions. At the beginning of the simulation, we initialize the user and item agent memories $M_u$ and $M_i$, using basic properties. Similar to most sequential recommendation settings~\cite{wang2019sequential, guo2023few}, for each user $u$, we use items $[i_1, i_2, \ldots, i_{n-1}]$, except the last item $i_n$ of the behavior sequences, as positive items for simulation. At each step of the interactions between $u$ and $i_j \in [i_1, i_2, \ldots, i_{n-1}]$, we randomly sample a negative item $i^-$ with high popularity among all items $I$ to help the user agent refine their profiles better through comparison. We use the last item $i_n$ as the ground-truth item in the ranking stage for testing.

For each user $u$ and item $i$, we first position these nodes in the KG. Then, we use our Path Extraction module to extract all 2-hop paths $P_{2}^{ui} = u \rightarrow e_1 \rightarrow i$ and 3-hop paths $P_{3}^{ui} = u \rightarrow e_1 \rightarrow e_2 \rightarrow i$ from the KG. Based on the extracted paths, we apply our Path Translator module to convert $P_{2}^{ui}$ to text $T_{2}^{ui}$ and $P_{3}^{ui}$ to text $T_{3}^{ui}$. Finally, we apply our Path Incorporation module to incorporate $T_{2}^{ui}$ and $T_{3}^{ui}$ into the LLM Agents' simulation and ranking process.

In the Autonomous Interaction phase, we ask the user agent to choose an item from $i^+$ and $i^-$ based on the current user and item memories $M_u$, $M_i$, and $M_{i^-}$, as well as the 2-hop relations between the user and items $T_2^{ui}$ and $T_2^{ui^-}$. After each interaction, the user agent selects an item $i_{select}$ and we also ask the user agent to provide explanations $y_{exp}$ for this choice. To optimize the user agent and item agents, we derive the feedback signal by comparing the user agent's chosen item with the actual interacted item. In the Reflection phase, we inform the user agent whether the previous choice was correct or incorrect. Simultaneously, we incorporate the knowledge graph information $T_{2}^{ui}$ and $T_{3}^{ui}$ to enable the user agents to analyze the reasons behind their choices abductively and summarize key factors to update their memory. We also update the memory of item agents based on the information from KG paths.

In the \textit{ranking stage}, we have user and item agents representing real-world user and item profiles. We focus on ranking the candidate items given a user $u$ and a set of candidates $\{c_1, ..., c_n\}$, the user agent's memory $M_u$, the memories of all candidates $\{M_{c1}, ..., M_{cn}\}$, and 2-hop KG information for each candidate. $R$ is the final ranking list. In the following sections, we will describe our detailed design to incorporate the KG information with LLM agents for recommendation. 

\subsection{Knowledge Graph Path Extraction}
This module aims to extract useful information from the knowledge graph $G$ for building better user and item profiles. 
In recommendation systems, given a user $u$ and his interaction history,  the critical information needed is the rationale behind why $u$ chooses or ignores item $i$. These rationales can provide detailed preference information that helps the user agent build a more accurate and comprehensive profile.
In $G$, paths between two nodes can explain why one node has a relation to another node, which can actually help explain why the user node $u$ chooses or does not choose the item node $i$ in $G$. Hence, our first task is to extract paths as additional contexts for the following prompt. To achieve this, we propose the following path extraction procedure: for each user $u$ and item $i$, we first extract the 2-hop knowledge path set $P_{2}^{ui}$ and 3-hop path set  $P_{3}^{ui}$, 
where $P_2$ contains all 2-hop paths and $P_3$ contains all 3-hop paths from $u$ to $i$.

\subsection{Path Translation: Translating 2-hop Paths to Text}

All 2-hop and 3-hop paths extracted from KG are represented as triples or quaternions. Since the input to LLM is in text format, our objective is to translate these 2-hop and 3-hop paths in a textual form that LLM can understand. We identify two critical challenges in this translation process: 
\textbf{Description Simplification} and \textbf{Easily comprehensible to LLM}.

Description Simplification refers to the need to simplify the textual representation of the extracted paths, as the number of 2-hop and 3-hop paths can be large. If we do not shorten the text, its length may exceed the token limit of the LLM. Furthermore, LLMs may struggle to capture important factors from longer contexts ~\cite{liu2024lost}.
Easily comprehensible to LLM refers to the need for presenting the path information in a way that the LLM can easily comprehend. While directly adding all 2-hop path triples to the prompt is one approach~\cite{shu2024knowledge}, this method necessitates introducing the entire KG to LLM, which is feasible only for very small graphs. Additionally, LLMs often struggle to effectively process and analyze these direct triples. 
In recommendation scenarios, we require the LLM to act as a reasoner, examining the potential critical reasons behind user choices based on the relationships between users and items. Thus, we aim to translate the triples into more natural, human-like language to improve the LLM agent's understanding of their underlying meaning.

 \Cref{alg:2hop} provides the details of translating 2-hop paths to text, and some examples of translating 2-hop and 3-hop to text are shown in \Cref{fig:framework}. We first group all paths for each user-item pair by the overall edge type of the paths. For each edge type $(r_1, r_2)$, we have a subset of paths associated with it. Since the start entity (user) and the end entity (item) are the same within each subset, we merge all paths by concatenating the second entities as a list. This approach reduces the number of 2-hop paths while still representing the relevant information. To make the text better understandable for the LLM, we describe the merged formulas by emphasizing the relationships between the user and the item. For example, given a merged paths set $u \xrightarrow{r_1} (e_1, e_2, ..., e_n) \xrightarrow{r_2} i$ with edge type $r_1=mentions, r_2=describe\_as$, we describe it as ``User mentions features $e_1, e_2, ..., e_n$  which are described as this item". This approach reduces the length of the 2-hop path information and explicitly prompts the LLM to perform better reasoning for why the user interacts with the item by highlighting the relationships.

\begin{algorithm}
\caption{\texttt{Trans-2HOP} ($G,u,i$)}\label{alg:2hop}
\begin{algorithmic}[1]
\Require \Statex Knowledge Graph $G$, user $u$, and item $i$
\Statex Function \texttt{FIND-2HOP}$(G, u, i)$, which retrieves all 2-hop paths between $u$ and $i$
\Ensure Text $T_2^{ui}$ containing natural language descriptions of 2-hop paths between $u$ and $i$
\State $P_2^{ui} \gets \texttt{FIND-2HOP}(G, u, i)$
\State Group paths by relation types $(r_1, r_2)$ in $P_2^{ui}$ to dictionary $D[(r_1, r_2)]$ where keys are relation pairs and values are lists of entities
\For{each $(r_1, r_2)$ in $D$}
    \State Formulate sentence: ``The user \{ $r_1$ \} $D[(r_1, r_2)]$ (list of entities), which are \{ $r_2$ \} by the item.''
    \State Append sentence to $T_2^{ui}$
\EndFor
\State \Return $T_2^{ui}$
\end{algorithmic}
\end{algorithm}

\subsection{Path Translation: Translating 3-hop Paths to Text}

\begin{algorithm}
\caption{
\texttt{Trans-3HOP} ($G,u,i,\{(u, i_1^+, i_1^-), \ldots, (u, i_n^+, i_n^-)\})$}
\label{alg:3hop}
\begin{algorithmic}[1]
\Require     \Statex Knowledge Graph $G$, user $u$, item $i$, and training samples for $u$: 
    \Statex $\{(u, i_1^+, i_1^-), \ldots, (u, i_n^+, i_n^-)\}$, where $i^+$ is a positive item and $i^-$ is a negative item sampled during autonomous interaction.
    \Statex Functions: \texttt{FIND-3HOP}$(G, u, i)$: retrieves all 3-hop paths between $u$ and $i$; \texttt{GET-DESC}$(P)$: returns entities (except $u$ and $i$) involved in path $P$
\Ensure Text $T_3^{ui}$ containing natural language descriptions of 3-hop paths between $u$ and $i$
\State Initialize: positive entity set $E^+ \gets \emptyset$, negative entity set $E^- \gets \emptyset$, non-informative entity set $S_u \gets \emptyset$
\For{each $(i_k^+, i_k^-)$ in training samples}
\State $E^+ \gets E^+ \cup \texttt{GET-DESC}(\texttt{FIND-3HOP}(G, u, i_k^+))$
    \State $E^- \gets E^- \cup \texttt{GET-DESC}(\texttt{FIND-3HOP}(G, u, i_k^-))$
\EndFor
\State $S_u \gets E^+ \cap E^-$ (identify non-informative common entities between positive and negative items).
\State $T_3^{ui} \gets \texttt{GET-DESC}(\texttt{FIND-3HOP}(G, u, i))$
\State Remove non-informative entities: $T_3^{ui} \gets T_3^{ui} \setminus S_u$
\State \Return $T_3^{ui}$
\end{algorithmic}
\end{algorithm}

Simplifying the description and making it simpler for the LLM to understand becomes more challenging for 3-hop paths because the number and relations of 3-hop paths are much larger compared to 2-hop paths. Unlike emphasizing the relations of 2-hop paths, the objective of using 3-hop paths is to incorporate more descriptive information for updating agents' memories in the Reflection stage.

\Cref{alg:3hop} provides the details of translating 3-hops to text. The number of 3-hop paths is too large, as shown in~\Cref{tab:exp_data_stat}. In the Autonomous Interaction and Reflection phase for the simulation stage, where the user needs to select one item among the positive and negative items, the motivation for adding KG information is to incorporate discriminative factors between these two items for the user agent. Therefore, we only need the discriminative features between the user's positive item pair and the user's negative item pair. Hence, we first construct a non-informative entity set $S_u$ for each user $u$. Specifically, entities that appear in both positive items and negative items indicate that they do not provide useful information for distinguishing user preferences, and thus they will be added to this set. Then, for user $u$ and item $i$, we extract the 3-hop paths from the KG, extract the descriptive entities, and filter based on $S_u$.  
The descriptive entities are pre-defined. For example, if a KG contains four types of entities: user $U$, item $I$, category $C$, and words $W$, we only select category $C$ and words $W$ as descriptive entities.
Finally, we obtain the filtered descriptive entities $T_3^{ui}$ which may indicate the user's potential rationales for their choice.

\subsection{Path Incorporation: Incorporating Text}

After obtaining 2-hop and 3-hop text descriptions from the KG $G$, we need to incorporate these descriptions into the overall framework. As shown in~\Cref{alg:framework}, we incorporate translated 2-hop KG text into the Autonomous Interaction stage (line 8), 2-hop and 3-hop KG text into the Reflection stage (lines 10 and 11), and 2-hop KG text into the Ranking stage (line 14). 




%% file: 4-Experiment.tex
\section{Experiments}

\subsection{Experiment Settings}

We conducted extensive experiments to address the following research questions (RQs):
\begin{itemize}
    \item {RQ1: Does incorporating KG information enhance the recommendation performance?}
    \item {RQ2: To what extent do different types of KG information enhance performance?}
    \item {RQ3: How does KG information influence agent
memory (user profiles) in simulation?}
    \item {RQ4: How does the enhanced agent's memory influence ranking?}
    \item {RQ5: How much does our method reduce the input word count for the LLM?}
\end{itemize}

\paragraph{Datasets.}
Following previous works, we conducted our experiments on three datasets containing KG: CDs, Clothing, and Beauty, which comprises product review data from existing recommendation benchmarking datasets~\cite{mcauley2015image}. The original data sets and Knowledge Graph (KG) information are sourced from~\cite{xian2019reinforcement}\footnote{\url{https://github.com/orcax/PGPR}}. To reduce the impact of expensive API calls and facilitate effective simulations, consistent with previous settings~\cite{zhang2024agentcf}, we sample dense recommendation subsets, where most users have rated a large portion of items. Specifically, we sample 100 users along with their corresponding items from each dataset for experiments. The full and sampled dataset statistics are presented in~\Cref{tab:exp_data_stat}. The Knowledge Graph comprises five types of entities:
\begin{itemize}
    \item User: User in recommender system
    \item Item: Product to be recommended to users
    \item Feature: A product feature word from reviews
    \item Brand: Brand or manufacturer of the product
    \item Category: Category of the product
\end{itemize}

and eight kinds of relations:
\begin{itemize}
    \item $User \xrightarrow{purchase} Item$
    \item $User \xrightarrow{mention} Feature$
    \item $Item \xrightarrow{describe\_as} Feature$
    \item $Item \xrightarrow{belong\_to} Category$
    \item $Item \xrightarrow{produced\_by} Brand$
    \item $Item \xrightarrow{also\_bought} another Item$
    \item $Item \xrightarrow{also\_viewed} another Item$
    \item $Item \xrightarrow{bought\_together\_by\_same\_user} another Item$
\end{itemize}

\begin{table}[h]
\caption{Statistics of the full, sampled datasets and the related KG information for the sampled datasets. \#Avg. 2-hop denotes the average number of 2-hop paths while \#Avg. 3-hop denotes the average number of 3-hop paths. }
\vspace{-10pt}

\centering
\resizebox{0.47\textwidth}{!}{
\begin{tabular}{llllll}
\hline
Datasets        & \#Users & \#Items & \#Interactions & \#Avg. 2-hop & \#Avg. 3-hop \\ \hline
CDs (full)      & 75258   & 64424   & 804090         & -            & -            \\
- Sample        & 100     & 247     & 506            & 13.59        & 349.77       \\ \hline
Clothing (full) & 39387   & 23030   & 214696         & -            & -            \\
- Sample        & 100     & 258     & 484            & 15.56        & 213.87       \\ \hline
Beauty (full)   & 22363   & 12100   & 149844         & -            & -            \\
- Sample        & 100     & 252     & 494            & 20.36        & 337.29       \\ \hline
\end{tabular}
}
\label{tab:exp_data_stat}
\vspace{-10pt}
\end{table}

\paragraph{Evaluation metrics.}
Similar to previous studies, we used NDCG@K as an evaluation metric for comparing different methods, where K is set to 1, 5, and 10. We considered the items excluding the last item of the user behavior sequences as the training data for the simulation stage in the recommendation, and the last item as the ground-truth item. To mitigate the randomness of the LLM, each experiment was repeated three times, and the mean and standard deviation were reported.

\begin{table*}[h]
\caption{The overall performance comparison in terms of NDCG metric. For the performance on our sampled datasets, we highlight the best result in bold font, and
the second-best result underlined.}
\vspace{-10pt}
\centering
\resizebox{\textwidth}{!}{
\begin{tabular}{llccccccccc}
\hline
\multicolumn{2}{l}{\multirow{2}{*}{Method}}                                                                        & \multicolumn{3}{c}{CDs}                                                                                                        & \multicolumn{3}{c}{Clothing}                                                                                                   & \multicolumn{3}{c}{Beauty}                                                                                                     \\ \cline{3-11} 
\multicolumn{2}{l}{}                                                                                               & NDCG@1                                   & NDCG@5                                   & NDCG@10                                  & NDCG@1                                   & NDCG@5                                   & NDCG@10                                  & NDCG@1                                   & NDCG@5                                   & NDCG@10                                  \\ \hline
\multicolumn{1}{l|}{\multirow{2}{*}{\begin{tabular}[c]{@{}l@{}}Trained on Full\\ Data\end{tabular}}} & $BPR_{full}$         & 0.300$\pm$0.014                          & 0.541$\pm$0.008                          & 0.617$\pm$0.002                          & 0.300$\pm$0.007                          & 0.496$\pm$0.020                          & 0.601$\pm$0.009                          & 0.420$\pm$0.014                          & 0.623$\pm$0.035                          & 0.688$\pm$0.017                          \\
\multicolumn{1}{l|}{}                                                                                & $SASRec_{full}$      & 0.590$\pm$0.007                          & 0.785$\pm$0.003                          & 0.807$\pm$0.004                          & 0.455$\pm$0.035                          & 0.675$\pm$0.019                          & 0.719$\pm$0.014                          & 0.675$\pm$0.007                          & 0.809$\pm$0.005                          & 0.835$\pm$0.000                          \\ \hline
\multicolumn{1}{l|}{\multirow{3}{*}{\begin{tabular}[c]{@{}l@{}}Conventional \\ Rec.\end{tabular}}}                     & BPR         & 0.083$\pm$0.021                          & 0.278$\pm$0.019                          & 0.441$\pm$0.018                          & 0.110$\pm$0.035                          & 0.307$\pm$0.021                          & 0.462$\pm$0.020                          & 0.113$\pm$0.031                          & 0.313$\pm$0.032                          & 0.468$\pm$0.021                          \\
\multicolumn{1}{l|}{}                                                                                & Pop         & 0.140$\pm$0.000                          & 0.346$\pm$0.000                          & 0.493$\pm$0.000                          & 0.050$\pm$0.000                          & 0.227$\pm$0.000                          & 0.407$\pm$0.000                          & 0.170$\pm$0.000                          & 0.359$\pm$0.000                          & 0.500$\pm$0.000                          \\
\multicolumn{1}{l|}{}                                                                                & BM25        & 0.050$\pm$0.000                          & 0.318$\pm$0.000                          & 0.451$\pm$0.000                          & 0.130$\pm$0.000                          & 0.333$\pm$0.000                          & 0.476$\pm$0.000                          & 0.180$\pm$0.000                          & 0.379$\pm$0.000                          & 0.523$\pm$0.000                          \\ \hline
\multicolumn{1}{l|}{Deep-learning Rec.}                                                              & SASRec      & 0.163$\pm$0.025                          & 0.346$\pm$0.019                          & 0.496$\pm$0.010                          & 0.157$\pm$0.015                          & 0.309$\pm$0.014                          & 0.481$\pm$0.009                          & 0.153$\pm$0.012                          & 0.352$\pm$0.010                          & 0.486$\pm$0.006                          \\ \hline
\multicolumn{1}{l|}{\multirow{3}{*}{\begin{tabular}[c]{@{}l@{}}LLM-based\\ Rec.\end{tabular}}}       & LLMRank     & 0.170$\pm$0.026                          & \underline{0.384$\pm$0.016} & \underline{0.515$\pm$0.012} & \underline{0.340$\pm$0.026} & \underline{0.648$\pm$0.016} & \underline{0.676$\pm$0.014} & 0.270$\pm$0.010                          & \underline{0.578$\pm$0.004} & \underline{0.624$\pm$0.005} \\
\multicolumn{1}{l|}{}                                                                                & AgentCF     & \underline{0.193$\pm$0.006} & 0.362$\pm$0.012                          & 0.510$\pm$0.006                          & 0.313$\pm$0.023                          & 0.528$\pm$0.020                          & 0.617$\pm$0.013                          & \underline{0.277$\pm$0.032} & 0.539$\pm$0.008                          & 0.607$\pm$0.017                          \\
\multicolumn{1}{l|}{}                                                                                & KGLA (Ours) & \textbf{0.377$\pm$0.006}                 & \textbf{0.637$\pm$0.009}                 & \textbf{0.675$\pm$0.005}                 & \textbf{0.453$\pm$0.021}                 & \textbf{0.699$\pm$0.016}                 & \textbf{0.732$\pm$0.010}                 & \textbf{0.390$\pm$0.010}                 & \textbf{0.655$\pm$0.003}                 & \textbf{0.691$\pm$0.003}                 \\ \hline
\multicolumn{2}{l}{Improvement over best baseline:}                                                                & 95.34\%                                  & 65.89\%                                  & 31.07\%                                  & 33.24\%                                  & 7.87\%                                   & 8.28\%                                   & 40.79\%                                  & 13.32\%                                  & 10.74\%                                  \\ \hline
\end{tabular}
}
\label{tab:exp_overall_performance}
\vspace{-10pt}
\end{table*}

\subsection{Baselines}
We compared the proposed model with the following categories of baseline methods available in the literature.

\paragraph{Conventional
recommendation methods.}
We included BPR (Bayesian Personalized Ranking)~\cite{rendle2012bpr} which uses matrix factorization to learn the potential representations of users and items and performs recommending based on these representations; Pop (Popularity-based recommendation)~\cite{ji2020re} which ranks candidates based on their popularity; and BM25 (Best Matching 25)~\cite{10.1561/1500000019} which ranks candidates based on their textual similarity to the user's past interactions.

\paragraph{Deep-learning based recommendation methods.}
We used SASRec~\cite{kang2018self} which captures the sequential patterns of the users' historical interactions utilizing a transformer-encoder.

\paragraph{LLM-based recommendation methods.}
We used LLMRank~\cite{hou2024large} as a baseline, which leverages LLMs as a zero-shot ranker to rank candidate items based on the sequential interaction history of the user. We compared our method with AgentCF~\cite{zhang2024agentcf} which also builds an Agent Simulation framework to obtain agent memories that are used for recommendation. Compared to their model, our method adaptively designs strategies to synergize the Agent with the KG.

\paragraph{Trained on Full Data}
Most of the above baselines are evaluated on sampled datasets to test their performance in recommendation scenarios within a few interactions. Additionally, We report the performance of BPR and SASRec (two widely used and strong baseline methods) which are trained on the full dataset, denoted $BPR_{full}$ and $SASRec_{full}$ respectively. It is important to note that these models are trained on the full dataset and serve as a reference to help examine the performance of our methods in recommendation scenarios with rich user-item interactions.

\paragraph{Implementation details.}
We implemented all baseline methods~\cite{hou2024large} using the open-source repository available at\footnote{\url{https://github.com/RUCAIBox/LLMRank}}.
For all Large Language Model (LLM)-based methods (LLMRank, AgentCF, and our methods), we employ Claude3-Haiku-20240307 as the LLM. We set the maximum number of tokens (max\_tokens) to 20,000. All other optional parameters are left at their default values: the temperature is 1, top\_p is 1, and top\_k is 250.

\subsection{RQ1: Does incorporating KG information enhance the recommendation performance?}
In this section, we conduct an in-depth analysis to investigate whether incorporating KG information can enhance recommendation performance across various settings, including different datasets, training dataset sizes, and LLMs. Compared to previous methods, our extensive experiments address the gap in exploring LLM-based agent methods under varying training dataset sizes and LLMs, providing valuable insights for the further development of LLM-based agents in recommendation systems.

\subsubsection{Overall Performance}
The overall performance of the recommendation measured by NDCG@K is reported in~\Cref{tab:exp_overall_performance}. From the results, we have the following observations:\\
\textbf{1)} Our method significantly outperforms all baselines in all datasets. Compared to the previous best baseline, our methods obtained 95.3\%, 44.7\%, and 40.8\% improvements in NDCG@1 on the three datasets. This shows that incorporating the KG information through our methods can significantly enhance the model's ability to make accurate recommendations.\\ 
\textbf{2)} The performance of all LLM-based recommendation methods surpasses that of conventional and deep learning methods. This indicates that LLMs can improve the recommendation performance, especially in settings with a small number of training datasets, due to their strong generalization capability. We will further investigate how will the superior performance of LLM-based methods change as the number of training datasets increases.\\
\textbf{3)} Compared to methods trained on the full dataset  ($BPR_{full}$ and $SASRec_{full}$), our method achieves comparable performance on Clothing dataset, while it performs less effectively on CDs and Beauty datasets. It is important to highlight $BPR_{full}$ and $SASRec_{full}$ are trained on full datasets, whereas our method is applied to interactions from only 100 users (approximately 0.06\%, 0.23\%, 0.33\% of full datasets for CDs, Clothing and Beauty, respectively). This demonstrates that, by leveraging LLMs and KGs, our method exhibits strong capabilities while preserving explainability in recommendation scenarios with limited interactions (Ex. System cold-start recommendation scenario~\cite{wu2023towards}). In scenarios with rich interactions, there remains a potential for improvement in LLM-based agent methods.

\subsubsection{Experiments with different number of users} 
\begin{figure*}[h]
\begin{center}
  \includegraphics[width=0.95\linewidth]{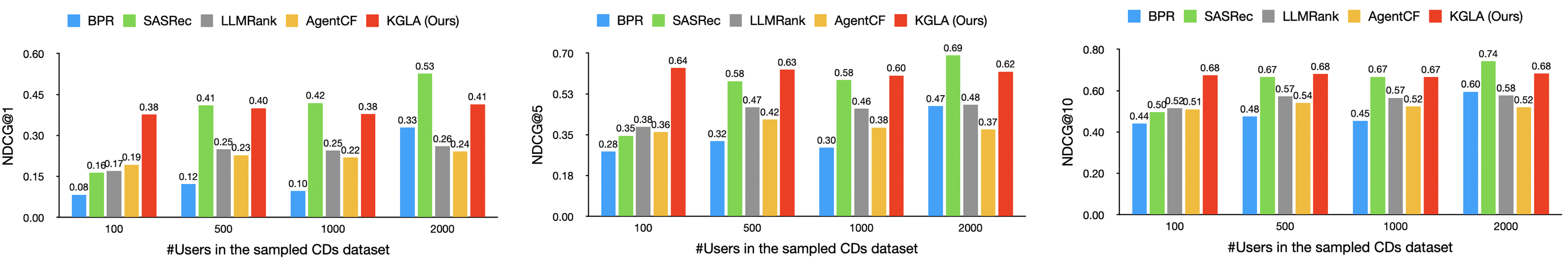}
\end{center}
\vspace{-5pt}
\caption{Experimental results across different numbers of users, with NDCG@1, NDCG@5, and NDCG@10 reported separately.}
\label{fig:different_users}
\vspace{-10pt}
\end{figure*}

Most previous LLM-based agent methods for recommendation primarily emphasize their superior few-shot capabilities on a small number of sampled datasets. In this work, we further investigate how this superior performance changes with the scale of the training dataset. As shown in Figure~\ref{fig:different_users}, we randomly sample the full CD datasets with 100, 500, 1000, and 2000 users to evaluate the performance of BPR, SASRec, and LLM-based methods (LLMRank, AgentCF, and our method). We observe the following: \\ 
\textbf{1)} In the 100-user setting, all LLM-based methods (LLMRank, AgentCF, and ours) outperform other methods, with our method achieving the best performance. This indicates that LLMs can serve as strong recommenders by leveraging their extensive pretrained knowledge. KGs can effectively supplement LLMs with rationales, as demonstrated by our method in few-user recommendation scenarios. \\
\textbf{2)} As the number of users increases, LLMRank and AgentCF perform worse than $SASRec_{full}$, while our method achieves comparable performance. This demonstrates that current LLM-based methods still struggle to effectively learn from large-scale interactions. The reason is that current LLM-based agent recommendation methods primarily rely on the specific user's history to build its profile and generate recommendations.
Consequently, increasing the number of other users has minimal direct impact on the profile of this specific user. \textit{Our work primarily focuses on integrating KG information into the LLM-based recommendation pipeline}. There remains significant potential for improving the scalability of LLM agents to learn effectively from large-scale interactions, which represents a promising direction for future LLM Multi-Agents research.

\subsubsection{Experiments using different LLMs}
As shown in Table~\ref{tab:exp_different_LLMs}, we evaluate different LLM-based methods using Claude-Haiku-20240307, LLama-3-8B-instruct, and GPT-4-turbo-2024-04-09. The observations are as follows:\\
\textbf{1)} Across all three LLMs, our method consistently outperforms other LLM-based methods, indicating that integrating KG information using our approach enhances recommendation performance and demonstrates the strong generalizability of our method. \\
\textbf{2)} GPT-4 achieves the best recommendation performance. This highlights that a larger amount of pretrained knowledge and stronger reasoning capabilities in more advanced LLM significantly contribute to a better inference of user preferences and improved recommendation performance.

\begin{table}[h]
\caption{The performance of LLM-based methods using different LLMs. The best results are in bold font.}
\vspace{-10pt}

\resizebox{0.47\textwidth}{!}{
\begin{tabular}{l|llll}
\hline
LLM                        & Method      & NDCG@1                                                        & NDCG@5                                                        & NDCG@10                                                       \\ \hline
                           & LLMRank     & 0.170                                                         & 0.384                                                         & 0.515                                                         \\
                           & AgentCF     & 0.193                                                         & 0.362                                                         & 0.510                                                         \\
\multirow{-3}{*}{Claude-Haiku}   & KGLA (Ours) & \textbf{0.377}                                                & \textbf{0.637}                                                & \textbf{0.675}                                                \\ \hline
                           & LLMRank     &  0.190          &  0.401         &  0.514          \\
                           & AgentCF     &  0.210          &  0.408          &  0.541          \\
\multirow{-3}{*}{LLama-3-8B} & KGLA (Ours) &  \textbf{0.350} &  \textbf{0.596} &  \textbf{0.654} \\ \hline
                           & LLMRank     & 0.200                                                         & 0.453                                                         & 0.545                                                         \\
                           & AgentCF     & 0.310                                                         & 0.611                                                         & 0.656                                                         \\
\multirow{-3}{*}{GPT-4-Turbo}    & KGLA (Ours) & \textbf{0.450}                                                & \textbf{0.669}                                                & \textbf{0.714}                                                \\ \hline
\end{tabular}
}
\label{tab:exp_different_LLMs}
\vspace{-10pt}
\end{table}







\subsection{RQ2: To what extent do different types of KG information enhance performance?}
After investigating the overall performance of our method, we further examined the impact of different types of KG information on recommendation performance through ablation studies on the CD dataset. The results are presented in~\Cref{tab:exp_ablation_study}, which show that incorporating 2-hop and 3-hop information from the KG can gradually improve the performance. This is because the average word count for 2-hop paths is 15.38, increasing to 62.73 for 3-hop paths (as shown in~\Cref{tab:exp_words_length}), suggesting that 3-hop paths provide richer information, which helps enhance the quality of user-profiles and improves performance. 
Our analysis indicates that while higher-hop paths offer richer information, the number of paths (word count) must be carefully reduced so that LLM can process them within their context length limitations. Exploring the integration of higher-hop paths from the KG for LLM presents a promising direction for future research.

\begin{table}[h]
\caption{The impact of 2-hop/3-hop KG paths information.}
\vspace{-10pt}
\resizebox{0.47\textwidth}{!}{
\begin{tabular}{llll}
\hline
Method                                                                   & NDCG@1                   & NDCG@5                   & NDCG@10                  \\ \hline
AgentCF (Baseline)                                                       & 0.193$\pm$0.006          & 0.362$\pm$0.012          & 0.510$\pm$0.006          \\ \hline
KGLA (+KG 2-hop)                                                         & 0.280$\pm$0.010           & 0.497$\pm$0.016          & 0.588$\pm$0.008          \\ \hline
KGLA (+KG 3-hop)                                                         & 0.257$\pm$0.029          & 0.518$\pm$0.011          & 0.596$\pm$0.012          \\ \hline
\begin{tabular}[c]{@{}l@{}}KGLA \\ (+KG 2-hop\\ + KG 3-hop)\end{tabular} & \textbf{0.377$\pm$0.006} & \textbf{0.637$\pm$0.009} & \textbf{0.675$\pm$0.005} \\ \hline
\end{tabular}
}
\label{tab:exp_ablation_study}
\vspace{-10pt}

\end{table}

\subsection{RQ3: How does KG information influence agent memory (user profiles) in simulation?}
Our proposed methods aim to ensure that the text is easily comprehensible for LLMs to analyze. In this experiment, to further investigate whether the LLM understands the incorporated text and how it contributes to building better user profiles, we conducted a case study to evaluate the impact of the KG information on the agent's memory during the Reflection stage. 

\begin{figure}[H]
\begin{center}
\includegraphics[scale=0.3]{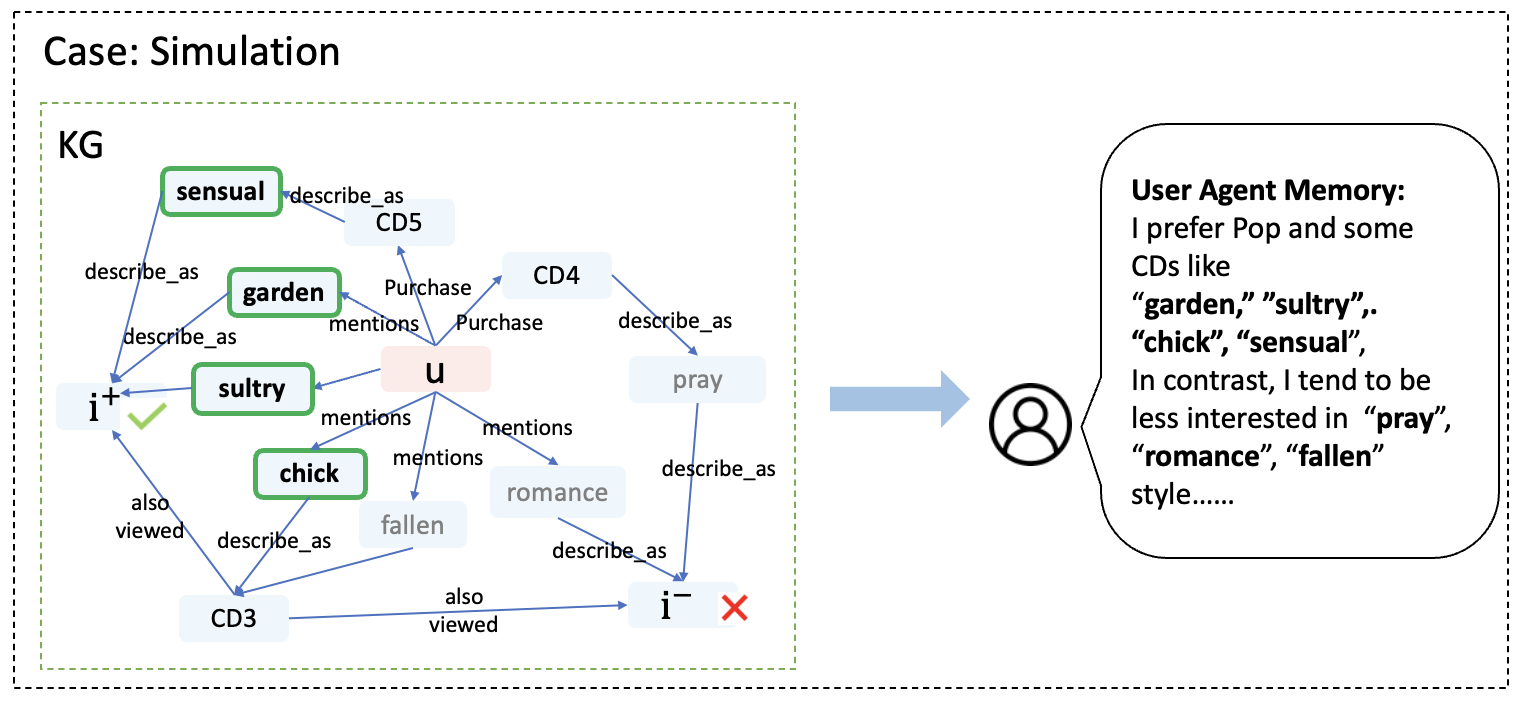}
\end{center}
\vspace{-10pt}
\caption{A case study for KG-enhanced simulation: Features in bold font such as ``sensual" and ``sultry" are summarized by the User Agents as the preferred features while light-colored features such as ``pray" and ``fallen" are summarized by User Agents as the non-informative features.} 
\label{fig:simulation}
\vspace{-10pt}
\end{figure}

To further investigate our method's main contribution to the agent's memory, we will provide a case study of our method with KG. As shown in ~\Cref{fig:simulation}, during simulation, with positive and negative items, the LLM can effectively conclude the potential preferred 2-hop features including ``garden", ``sultry", ``chick" related to the user and help explore the expanded 3-hop features including ``sensual" to enrich the user profiles.

\begin{figure}[H]
\begin{center}
\includegraphics[scale=0.28]{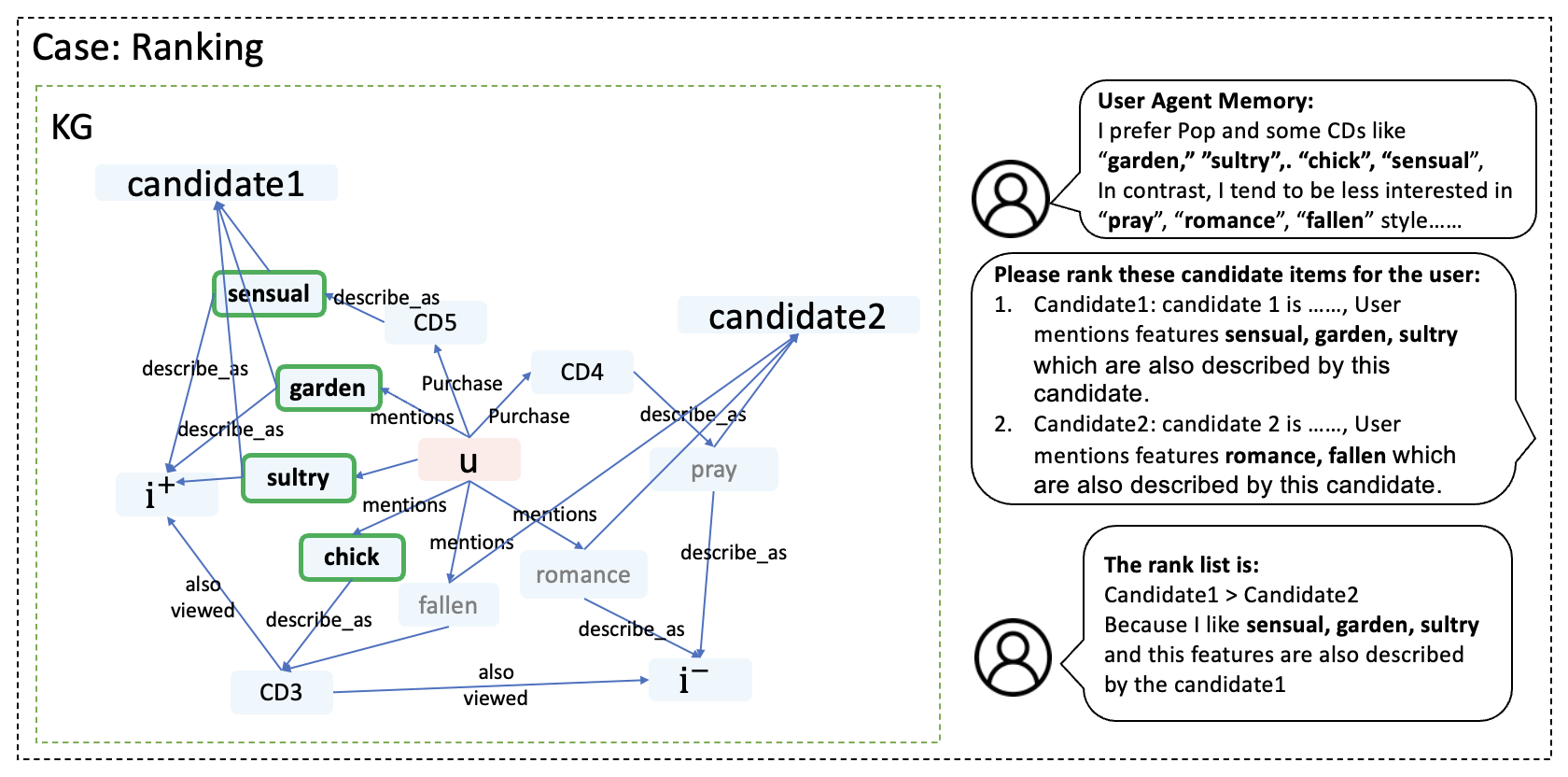}
\end{center}
\vspace{-10pt}
\caption{A case study for KG-enhanced ranking: Taking two item candidates as examples, Candidate 1 aligns more closely with the user agent's memory, enabling the user agent to make the correct choice during the ranking stage.}
\label{fig:ranking}
\vspace{-5pt}
\end{figure}

\begin{table*}[ht]
\caption{An example showing the effect of KG information in simulation and ranking for recommendation.}
\vspace{-10pt}

\centering
\scriptsize
\begin{tabular}{|p{2\columnwidth}|}
\hline
\textbf{1. Initialization} \\ \hline
\textbf{User Agent:} I enjoy listening to CDs very much. \\
\textbf{Positive Item Agent: } The CD is called “Brainwashed”. The category of this CD is: “Classic Rock; Album-Oriented Rock (AOR)”. \\
\textbf{Negative Item Agent: } The CD is called “O, Yeah! Ultimate Aerosmith Hits”. The category of this CD is: “Classic Rock \dots. \\ \hline
\textbf{2. Autonomous Interaction} \\ \hline
\textbf{System Prompt for User Agent (KG information is highlighted in blue, with blue parts omitted for baseline without KG.):} \\ 
You are a CD enthusiast. Here is your self-introduction \dots \{user agent memory\}. Now, you are selecting a CD from: CD1: The CD is called [Justified]. The category of this CD is: Teen Pop \dots . \textcolor{blue}{The relations between you and this CD is You mentions `seat' `garden' \dots `parade', and these features also described by this CD.}
CD2: The CD is called [Metallica]. The category of this CD is: Metal \dots \textcolor{blue}{The relations between you and this CD is You mentions `empty' `suprise' `pray' `internet' \dots `tap', and these features also described  by this CD.}. Please select the CD that aligns best with your preferences. \dots  \\ \hline
\end{tabular}

\begin{tabular}{|p{0.58\columnwidth}|p{1.38\columnwidth}|}
\hline
\textbf{User Agent response Without KG:} &
\textbf{User Agent response With KG (Blue parts show the Agent justify the preference grounded by KG factual information):} \\
Chosen CD: Metallica
Explanation: Based on the self-introduction provided, I have a strong preference for CDs that seamlessly integrate diverse yet complementary musical elements \dots  &
Chosen CD: Justified. Explanation:
As a CD enthusiast, my preferences and dislikes revealed in the self-introduction are as follows:
When evaluating the two candidate CDs, ``Justified" aligns much more closely with my current preferences compared to ``Metallica": Justified falls under the categories of Teen Pop  \dots \textcolor{blue}{The relations between the CD and myself, such as ``seat," ``garden," \dots  suggest a more introspective and emotionally resonant listening experience.}

In contrast, Metallica is firmly rooted in the Metal, Pop Metal, Thrash \& Speed Metal, and Hard Rock genres, \textcolor{blue}{which directly contradict my stated dislike \dots . The relations between the CD and myself, such as \dots  ``pray," and ``rhythm," do not indicate the same level of personal connection and emotional depth \dots}. Therefore, I have chosen the ``Justified" CD \dots \\
\hline
\end{tabular}

\begin{tabular}{|p{2\columnwidth}|}
\hline
\textbf{3. Reflection} \\ \hline
\end{tabular}

\begin{tabular}{|p{0.58\columnwidth}|p{1.38\columnwidth}|}
\hline
\textbf{System Prompt for User Agent Without KG} & \textbf{System Prompt for User Agent With KG (KG information is highlighted in blue)}  \\
You are a CD enthusiast.  \dots  Recently, you considered to choose one CD from two candidates. The features of these two candidate CDs are listed as follows: 1. \{neg item memory\}  2. \{pos item memory\} .
    
    After comparing these two candidates you selected \{neg item title\} to listen to while rejected the other one. \dots 
    However you discovered that you don’t like the CD that you initially chose \dots Therefore, your task now is to update your self-introduction \dots  &
You are a CD enthusiast. \dots Recently, you considered to choose one CD from two candidates. 
The features of these two candidate CDs are listed as follows:"CD1: \{neg item memory\}. \textcolor{blue}{The relations between you and this CD is You mentions `seat' `garden' \dots `parade', and these features also described by this CD.}
CD2: \{pos item memory\}. \textcolor{blue}{The relations between you and this CD is you mentions `empty' \dots `tap', and these features also described by this CD."}
\dots you made a correct choice \dots  your task now is to update your self-introduction \dots  When summarizing your preference from Justified, you can refer to these features which are highly related to Justified: \textcolor{blue}{ `songstress', `popiest', `quotscreamsquot', \dots `convincing', `sassy', , ``}.

When summarizing your dislikes from Metallica, you can refer to these features which are highly related to your updated dislikes: \textcolor{blue}{ `highest', `itit', `bow', \dots, `fortunate', `corporate', `tad'. 
}  \\ \hline
\textbf{User Agent Memory Update Without KG}: &  \textbf{User Agent Memory Update With KG}: \\
Response: My updated self-introduction: As a CD enthusiast, I have developed a strong appreciation for CDs that seamlessly blend diverse genres, such as the combination of teen pop, rock, pop, and R\&B found in ``Justified \dots &
Response: My updated self-introduction: As a CD enthusiast, I am particularly drawn to singer-songwriter and vocal pop genres, such as the ``Justified" CD \dots
 \textcolor{blue}{. The ``popiest," \dots
  ``sensual," and ``sultry" qualities of this album captivate me, as they provide a listening experience that is deeply engaging.} \dots  \textcolor{blue}{The ``Metallica" CD, with its ``highest," ``exotic," and ``supernatural" undertones, fails to resonate with me as it lacks the nuanced and introspective qualities I now seek in my music.}
\\ \hline
\end{tabular}
\begin{tabular}{|p{2\columnwidth}|}
\hline
\textbf{4. Ranking} \\ \hline
\textbf{System Prompt for Ranking Based on User Agent Memory:}\\
You are a CD enthusiast. Here is your past preferences and dislikes:\{user agent memory\}. Please rank these CDs by measuring the possibilities that the CD aligns best with your preferences. 
\\
1. The CD is called [Measure of a Man]. The category of this CD is: Dance Pop; Teen Pop; \dots." \textcolor{blue}{You mentions `afternoon' `garden' `suprise' \dots `carey' `parade', and these features also described  by this CD.} \\ 
\dots \\
10. The CD is called [Back in Black]. The category of this CD is: Metal; Arena Rock \dots. \textcolor{blue}{You mentions `empty' `earn' `rythm' `stood' `tap', and these features also described by this CD.}\\
\hline
\end{tabular}
\label{tab:details}
\end{table*}

\subsection{RQ4: How does the enhanced agent's memory influence the ranking?}
To investigate how the updated agent memory based on KG information influences the final recommendation, we present the following case study during the recommendation stage. As illustrated in~\Cref{fig:ranking}, with the precise and extensive user profiles and 2-hop relations, Candidate 1 shares more common features with user agent memory, thus providing the user agent with more rationales to rank the candidates.

\subsection{RQ5: How much does our method reduce the input word count for the LLM?}
A key objective of our proposed method is to shorten the description of the KG. In this experiment, we compare the word count of the original paths with the text generated by our methods for 2-hop and 3-hop paths. As shown in~\Cref{tab:exp_words_length}, our methods achieve a substantial reduction in word count, with approximately 60\% for 2-hop paths and 98\% for 3-hop paths, compared to the original descriptions of the path information across all datasets.
\begin{table}[h]
\caption{Reduction in word counts using the proposed method. `\# Avg. original w.' denotes the word count when the paths are directly put into the LLM. `\# Avg. words' denotes the word count after processing by our method.}
\vspace{-10pt}
\resizebox{0.45\textwidth}{!}{
\begin{tabular}{lllll}
\hline
\multicolumn{2}{l}{Word Count}                                      & \multicolumn{1}{c}{CDs} & Clothing & Beauty  \\ \hline
\multicolumn{1}{l|}{\multirow{4}{*}{2-hop Paths}} & \# Avg. paths        & 13.38                   & 15.56    & 20.36   \\
\multicolumn{1}{l|}{}                             & \# Avg. original w.  & 40.14                   & 46.68    & 61.08   \\
\multicolumn{1}{l|}{}                             & \# Avg. words        & 15.38                   & 17.56    & 22.36   \\
\multicolumn{1}{l|}{}                             & Reduction Percentage & 61.68\%                 & 62.38\%  & 63.39\% \\ \hline
\multicolumn{1}{l|}{\multirow{4}{*}{3-hop paths}} & \# Avg. paths        & 312.89                  & 213.87   & 337.29  \\
\multicolumn{1}{l|}{}                             & \# Avg. original w.  & 1564.45                 & 1069.35  & 1686.45 \\
\multicolumn{1}{l|}{}                             & \# Avg. words        & 62.73                   & 18.23    & 25.46   \\
\multicolumn{1}{l|}{}                             & Reduction Percentage & 95.99\%                 & 98.30\%  & 98.49\% \\ \hline
\end{tabular}
}
\label{tab:exp_words_length}
\vspace{-10pt}
\end{table}







\subsection{The Effect of KG Information throughout the Simulation}
\label{sec:effect_agent_memory}
In this section, we present key logs from the simulation process both without KG~\cite{zhang2024agentcf} and with KG using our method for comparison, as shown in~\Cref{tab:details}. KG-related prompts and responses throughout the process are highlighted in blue. Due to page limitations, some of the less relevant prompts and responses have been replaced with ellipses.

%% file: 5-Conclusion.tex
\section{Conclusion}

Using language agents for recommendation is a promising yet challenging task. Previous works on this topic have primarily focused on utilizing agents to simulate the recommendation process, often neglecting the underlying rationale behind recommendations. As a result, they have struggled to discover user preferences for recommendations.
In this paper, we propose KGLA—the first framework that explores the synergistic combination of LLM agents and KGs, enabling agents to act with improved rationale and better align their preferences with actual recommendations.
By leveraging existing datasets, we conducted extensive experiments to demonstrate how incorporating KG information refines the agent's memory with precise rationales, enabling our method to significantly outperform other baselines. Furthermore, our KG-enhanced framework assists agents in reflecting on grounding faithful KG information and provides better explanations of the recommendation process.
For practical applications, we identify three key uses for our method: 1) In recommendation scenarios with limited interactions, our method outperforms other approaches. 2) The text profiles generated by our method can supplement existing deep learning-based recommendation systems. 3) By explicitly modeling user preferences through text, our method offers explainable recommendations, allowing users to interact with their text profiles for enhanced interactive recommendations.

%% file: arxiv_sigconf.bbl

\begin{thebibliography}{32}


\ifx \showCODEN    \undefined \def \showCODEN     #1{\unskip}     \fi
\ifx \showISBNx    \undefined \def \showISBNx     #1{\unskip}     \fi
\ifx \showISBNxiii \undefined \def \showISBNxiii  #1{\unskip}     \fi
\ifx \showISSN     \undefined \def \showISSN      #1{\unskip}     \fi
\ifx \showLCCN     \undefined \def \showLCCN      #1{\unskip}     \fi
\ifx \shownote     \undefined \def \shownote      #1{#1}          \fi
\ifx \showarticletitle \undefined \def \showarticletitle #1{#1}   \fi
\ifx \showURL      \undefined \def \showURL       {\relax}        \fi
\providecommand\bibfield[2]{#2}
\providecommand\bibinfo[2]{#2}
\providecommand\natexlab[1]{#1}
\providecommand\showeprint[2][]{arXiv:#2}

\bibitem[Guo et~al\mbox{.}(2024)]%
        {guo2024large}
\bibfield{author}{\bibinfo{person}{Taicheng Guo}, \bibinfo{person}{Xiuying Chen}, \bibinfo{person}{Yaqi Wang}, \bibinfo{person}{Ruidi Chang}, \bibinfo{person}{Shichao Pei}, \bibinfo{person}{Nitesh~V Chawla}, \bibinfo{person}{Olaf Wiest}, {and} \bibinfo{person}{Xiangliang Zhang}.} \bibinfo{year}{2024}\natexlab{}.
\newblock \showarticletitle{Large language model based multi-agents: A survey of progress and challenges}.
\newblock \bibinfo{journal}{\emph{arXiv preprint arXiv:2402.01680}} (\bibinfo{year}{2024}).
\newblock


\bibitem[Guo et~al\mbox{.}(2023)]%
        {guo2023few}
\bibfield{author}{\bibinfo{person}{Taicheng Guo}, \bibinfo{person}{Lu Yu}, \bibinfo{person}{Basem Shihada}, {and} \bibinfo{person}{Xiangliang Zhang}.} \bibinfo{year}{2023}\natexlab{}.
\newblock \showarticletitle{Few-shot news recommendation via cross-lingual transfer}. In \bibinfo{booktitle}{\emph{Proceedings of the ACM Web Conference 2023}}. \bibinfo{pages}{1130--1140}.
\newblock


\bibitem[Hou et~al\mbox{.}(2024)]%
        {hou2024large}
\bibfield{author}{\bibinfo{person}{Yupeng Hou}, \bibinfo{person}{Junjie Zhang}, \bibinfo{person}{Zihan Lin}, \bibinfo{person}{Hongyu Lu}, \bibinfo{person}{Ruobing Xie}, \bibinfo{person}{Julian McAuley}, {and} \bibinfo{person}{Wayne~Xin Zhao}.} \bibinfo{year}{2024}\natexlab{}.
\newblock \showarticletitle{Large language models are zero-shot rankers for recommender systems}. In \bibinfo{booktitle}{\emph{European Conference on Information Retrieval}}. Springer, \bibinfo{pages}{364--381}.
\newblock


\bibitem[Huang et~al\mbox{.}(2023)]%
        {huang2023recommender}
\bibfield{author}{\bibinfo{person}{Xu Huang}, \bibinfo{person}{Jianxun Lian}, \bibinfo{person}{Yuxuan Lei}, \bibinfo{person}{Jing Yao}, \bibinfo{person}{Defu Lian}, {and} \bibinfo{person}{Xing Xie}.} \bibinfo{year}{2023}\natexlab{}.
\newblock \showarticletitle{Recommender ai agent: Integrating large language models for interactive recommendations}.
\newblock \bibinfo{journal}{\emph{arXiv preprint arXiv:2308.16505}} (\bibinfo{year}{2023}).
\newblock


\bibitem[Ji et~al\mbox{.}(2020)]%
        {ji2020re}
\bibfield{author}{\bibinfo{person}{Yitong Ji}, \bibinfo{person}{Aixin Sun}, \bibinfo{person}{Jie Zhang}, {and} \bibinfo{person}{Chenliang Li}.} \bibinfo{year}{2020}\natexlab{}.
\newblock \showarticletitle{A re-visit of the popularity baseline in recommender systems}. In \bibinfo{booktitle}{\emph{Proceedings of the 43rd International ACM SIGIR Conference on Research and Development in Information Retrieval}}. \bibinfo{pages}{1749--1752}.
\newblock


\bibitem[Jiang et~al\mbox{.}(2024)]%
        {jiang2024kg}
\bibfield{author}{\bibinfo{person}{Jinhao Jiang}, \bibinfo{person}{Kun Zhou}, \bibinfo{person}{Wayne~Xin Zhao}, \bibinfo{person}{Yang Song}, \bibinfo{person}{Chen Zhu}, \bibinfo{person}{Hengshu Zhu}, {and} \bibinfo{person}{Ji-Rong Wen}.} \bibinfo{year}{2024}\natexlab{}.
\newblock \showarticletitle{Kg-agent: An efficient autonomous agent framework for complex reasoning over knowledge graph}.
\newblock \bibinfo{journal}{\emph{arXiv preprint arXiv:2402.11163}} (\bibinfo{year}{2024}).
\newblock


\bibitem[Kang and McAuley(2018)]%
        {kang2018self}
\bibfield{author}{\bibinfo{person}{Wang-Cheng Kang} {and} \bibinfo{person}{Julian McAuley}.} \bibinfo{year}{2018}\natexlab{}.
\newblock \showarticletitle{Self-attentive sequential recommendation}. In \bibinfo{booktitle}{\emph{2018 IEEE international conference on data mining (ICDM)}}. IEEE, \bibinfo{pages}{197--206}.
\newblock


\bibitem[Kemper et~al\mbox{.}(2024)]%
        {kemper2024retrieval}
\bibfield{author}{\bibinfo{person}{Sara Kemper}, \bibinfo{person}{Justin Cui}, \bibinfo{person}{Kai Dicarlantonio}, \bibinfo{person}{Kathy Lin}, \bibinfo{person}{Danjie Tang}, \bibinfo{person}{Anton Korikov}, {and} \bibinfo{person}{Scott Sanner}.} \bibinfo{year}{2024}\natexlab{}.
\newblock \showarticletitle{Retrieval-Augmented Conversational Recommendation with Prompt-based Semi-Structured Natural Language State Tracking}. In \bibinfo{booktitle}{\emph{Proceedings of the 47th International ACM SIGIR Conference on Research and Development in Information Retrieval}}. \bibinfo{pages}{2786--2790}.
\newblock


\bibitem[Liao et~al\mbox{.}(2024)]%
        {10.1145/3626772.3657690}
\bibfield{author}{\bibinfo{person}{Jiayi Liao}, \bibinfo{person}{Sihang Li}, \bibinfo{person}{Zhengyi Yang}, \bibinfo{person}{Jiancan Wu}, \bibinfo{person}{Yancheng Yuan}, \bibinfo{person}{Xiang Wang}, {and} \bibinfo{person}{Xiangnan He}.} \bibinfo{year}{2024}\natexlab{}.
\newblock \showarticletitle{LLaRA: Large Language-Recommendation Assistant}. In \bibinfo{booktitle}{\emph{Proceedings of the 47th International ACM SIGIR Conference on Research and Development in Information Retrieval}} (Washington DC, USA) \emph{(\bibinfo{series}{SIGIR '24})}. \bibinfo{publisher}{Association for Computing Machinery}, \bibinfo{address}{New York, NY, USA}, \bibinfo{pages}{1785–1795}.
\newblock
\showISBNx{9798400704314}
\href{https://doi.org/10.1145/3626772.3657690}{doi:\nolinkurl{10.1145/3626772.3657690}}


\bibitem[Liu et~al\mbox{.}(2024)]%
        {liu2024lost}
\bibfield{author}{\bibinfo{person}{Nelson~F Liu}, \bibinfo{person}{Kevin Lin}, \bibinfo{person}{John Hewitt}, \bibinfo{person}{Ashwin Paranjape}, \bibinfo{person}{Michele Bevilacqua}, \bibinfo{person}{Fabio Petroni}, {and} \bibinfo{person}{Percy Liang}.} \bibinfo{year}{2024}\natexlab{}.
\newblock \showarticletitle{Lost in the middle: How language models use long contexts}.
\newblock \bibinfo{journal}{\emph{Transactions of the Association for Computational Linguistics}}  \bibinfo{volume}{12} (\bibinfo{year}{2024}), \bibinfo{pages}{157--173}.
\newblock


\bibitem[Luo et~al\mbox{.}(2023)]%
        {luo2023reasoning}
\bibfield{author}{\bibinfo{person}{Linhao Luo}, \bibinfo{person}{Yuan-Fang Li}, \bibinfo{person}{Gholamreza Haffari}, {and} \bibinfo{person}{Shirui Pan}.} \bibinfo{year}{2023}\natexlab{}.
\newblock \showarticletitle{Reasoning on graphs: Faithful and interpretable large language model reasoning}.
\newblock \bibinfo{journal}{\emph{arXiv preprint arXiv:2310.01061}} (\bibinfo{year}{2023}).
\newblock


\bibitem[McAuley et~al\mbox{.}(2015)]%
        {mcauley2015image}
\bibfield{author}{\bibinfo{person}{Julian McAuley}, \bibinfo{person}{Christopher Targett}, \bibinfo{person}{Qinfeng Shi}, {and} \bibinfo{person}{Anton Van Den~Hengel}.} \bibinfo{year}{2015}\natexlab{}.
\newblock \showarticletitle{Image-based recommendations on styles and substitutes}. In \bibinfo{booktitle}{\emph{Proceedings of the 38th international ACM SIGIR conference on research and development in information retrieval}}. \bibinfo{pages}{43--52}.
\newblock


\bibitem[Rendle et~al\mbox{.}(2012)]%
        {rendle2012bpr}
\bibfield{author}{\bibinfo{person}{Steffen Rendle}, \bibinfo{person}{Christoph Freudenthaler}, \bibinfo{person}{Zeno Gantner}, {and} \bibinfo{person}{Lars Schmidt-Thieme}.} \bibinfo{year}{2012}\natexlab{}.
\newblock \showarticletitle{BPR: Bayesian personalized ranking from implicit feedback}.
\newblock \bibinfo{journal}{\emph{arXiv preprint arXiv:1205.2618}} (\bibinfo{year}{2012}).
\newblock


\bibitem[Robertson and Zaragoza(2009)]%
        {10.1561/1500000019}
\bibfield{author}{\bibinfo{person}{Stephen Robertson} {and} \bibinfo{person}{Hugo Zaragoza}.} \bibinfo{year}{2009}\natexlab{}.
\newblock \showarticletitle{The Probabilistic Relevance Framework: BM25 and Beyond}.
\newblock \bibinfo{journal}{\emph{Found. Trends Inf. Retr.}} \bibinfo{volume}{3}, \bibinfo{number}{4} (\bibinfo{date}{apr} \bibinfo{year}{2009}), \bibinfo{pages}{333–389}.
\newblock
\showISSN{1554-0669}
\href{https://doi.org/10.1561/1500000019}{doi:\nolinkurl{10.1561/1500000019}}


\bibitem[Shi et~al\mbox{.}(2024)]%
        {shi2024enhancing}
\bibfield{author}{\bibinfo{person}{Wentao Shi}, \bibinfo{person}{Xiangnan He}, \bibinfo{person}{Yang Zhang}, \bibinfo{person}{Chongming Gao}, \bibinfo{person}{Xinyue Li}, \bibinfo{person}{Jizhi Zhang}, \bibinfo{person}{Qifan Wang}, {and} \bibinfo{person}{Fuli Feng}.} \bibinfo{year}{2024}\natexlab{}.
\newblock \showarticletitle{Enhancing Long-Term Recommendation with Bi-level Learnable Large Language Model Planning}.
\newblock \bibinfo{journal}{\emph{arXiv preprint arXiv:2403.00843}} (\bibinfo{year}{2024}).
\newblock


\bibitem[Shinn et~al\mbox{.}(2023)]%
        {shinn2023reflexion}
\bibfield{author}{\bibinfo{person}{Noah Shinn}, \bibinfo{person}{Beck Labash}, {and} \bibinfo{person}{Ashwin Gopinath}.} \bibinfo{year}{2023}\natexlab{}.
\newblock \showarticletitle{Reflexion: an autonomous agent with dynamic memory and self-reflection}.
\newblock \bibinfo{journal}{\emph{arXiv preprint arXiv:2303.11366}} \bibinfo{volume}{2}, \bibinfo{number}{5} (\bibinfo{year}{2023}), \bibinfo{pages}{9}.
\newblock


\bibitem[Shu et~al\mbox{.}(2024)]%
        {shu2024knowledge}
\bibfield{author}{\bibinfo{person}{Dong Shu}, \bibinfo{person}{Tianle Chen}, \bibinfo{person}{Mingyu Jin}, \bibinfo{person}{Yiting Zhang}, \bibinfo{person}{Mengnan Du}, {and} \bibinfo{person}{Yongfeng Zhang}.} \bibinfo{year}{2024}\natexlab{}.
\newblock \showarticletitle{Knowledge Graph Large Language Model (KG-LLM) for Link Prediction}.
\newblock \bibinfo{journal}{\emph{arXiv preprint arXiv:2403.07311}} (\bibinfo{year}{2024}).
\newblock


\bibitem[Shu et~al\mbox{.}(2023)]%
        {shu2023rah}
\bibfield{author}{\bibinfo{person}{Yubo Shu}, \bibinfo{person}{Hansu Gu}, \bibinfo{person}{Peng Zhang}, \bibinfo{person}{Haonan Zhang}, \bibinfo{person}{Tun Lu}, \bibinfo{person}{Dongsheng Li}, {and} \bibinfo{person}{Ning Gu}.} \bibinfo{year}{2023}\natexlab{}.
\newblock \showarticletitle{Rah! recsys-assistant-human: A human-central recommendation framework with large language models}.
\newblock \bibinfo{journal}{\emph{arXiv preprint arXiv:2308.09904}} (\bibinfo{year}{2023}).
\newblock


\bibitem[Sun et~al\mbox{.}(2024)]%
        {sun2024large}
\bibfield{author}{\bibinfo{person}{Zhu Sun}, \bibinfo{person}{Hongyang Liu}, \bibinfo{person}{Xinghua Qu}, \bibinfo{person}{Kaidong Feng}, \bibinfo{person}{Yan Wang}, {and} \bibinfo{person}{Yew~Soon Ong}.} \bibinfo{year}{2024}\natexlab{}.
\newblock \showarticletitle{Large language models for intent-driven session recommendations}. In \bibinfo{booktitle}{\emph{Proceedings of the 47th International ACM SIGIR Conference on Research and Development in Information Retrieval}}. \bibinfo{pages}{324--334}.
\newblock


\bibitem[Wang et~al\mbox{.}(2023b)]%
        {wang2023user}
\bibfield{author}{\bibinfo{person}{Lei Wang}, \bibinfo{person}{Jingsen Zhang}, \bibinfo{person}{Hao Yang}, \bibinfo{person}{Zhiyuan Chen}, \bibinfo{person}{Jiakai Tang}, \bibinfo{person}{Zeyu Zhang}, \bibinfo{person}{Xu Chen}, \bibinfo{person}{Yankai Lin}, \bibinfo{person}{Ruihua Song}, \bibinfo{person}{Wayne~Xin Zhao}, {et~al\mbox{.}}} \bibinfo{year}{2023}\natexlab{b}.
\newblock \showarticletitle{User behavior simulation with large language model based agents}.
\newblock \bibinfo{journal}{\emph{arXiv preprint arXiv:2306.02552}} (\bibinfo{year}{2023}).
\newblock


\bibitem[Wang et~al\mbox{.}(2019)]%
        {wang2019sequential}
\bibfield{author}{\bibinfo{person}{Shoujin Wang}, \bibinfo{person}{Liang Hu}, \bibinfo{person}{Yan Wang}, \bibinfo{person}{Longbing Cao}, \bibinfo{person}{Quan~Z Sheng}, {and} \bibinfo{person}{Mehmet Orgun}.} \bibinfo{year}{2019}\natexlab{}.
\newblock \showarticletitle{Sequential recommender systems: challenges, progress and prospects}.
\newblock \bibinfo{journal}{\emph{arXiv preprint arXiv:2001.04830}} (\bibinfo{year}{2019}).
\newblock


\bibitem[Wang et~al\mbox{.}(2023a)]%
        {wang2023recmind}
\bibfield{author}{\bibinfo{person}{Yancheng Wang}, \bibinfo{person}{Ziyan Jiang}, \bibinfo{person}{Zheng Chen}, \bibinfo{person}{Fan Yang}, \bibinfo{person}{Yingxue Zhou}, \bibinfo{person}{Eunah Cho}, \bibinfo{person}{Xing Fan}, \bibinfo{person}{Xiaojiang Huang}, \bibinfo{person}{Yanbin Lu}, {and} \bibinfo{person}{Yingzhen Yang}.} \bibinfo{year}{2023}\natexlab{a}.
\newblock \showarticletitle{Recmind: Large language model powered agent for recommendation}.
\newblock \bibinfo{journal}{\emph{arXiv preprint arXiv:2308.14296}} (\bibinfo{year}{2023}).
\newblock


\bibitem[Wang et~al\mbox{.}(2024)]%
        {wang2024multi}
\bibfield{author}{\bibinfo{person}{Zhefan Wang}, \bibinfo{person}{Yuanqing Yu}, \bibinfo{person}{Wendi Zheng}, \bibinfo{person}{Weizhi Ma}, {and} \bibinfo{person}{Min Zhang}.} \bibinfo{year}{2024}\natexlab{}.
\newblock \showarticletitle{Multi-Agent Collaboration Framework for Recommender Systems}.
\newblock \bibinfo{journal}{\emph{arXiv preprint arXiv:2402.15235}} (\bibinfo{year}{2024}).
\newblock


\bibitem[Wu et~al\mbox{.}(2023)]%
        {wu2023towards}
\bibfield{author}{\bibinfo{person}{Xuansheng Wu}, \bibinfo{person}{Huachi Zhou}, \bibinfo{person}{Wenlin Yao}, \bibinfo{person}{Xiao Huang}, {and} \bibinfo{person}{Ninghao Liu}.} \bibinfo{year}{2023}\natexlab{}.
\newblock \showarticletitle{Towards personalized cold-start recommendation with prompts}.
\newblock \bibinfo{journal}{\emph{arXiv preprint arXiv:2306.17256}} (\bibinfo{year}{2023}).
\newblock


\bibitem[Xi et~al\mbox{.}(2023)]%
        {xi2023rise}
\bibfield{author}{\bibinfo{person}{Zhiheng Xi}, \bibinfo{person}{Wenxiang Chen}, \bibinfo{person}{Xin Guo}, \bibinfo{person}{Wei He}, \bibinfo{person}{Yiwen Ding}, \bibinfo{person}{Boyang Hong}, \bibinfo{person}{Ming Zhang}, \bibinfo{person}{Junzhe Wang}, \bibinfo{person}{Senjie Jin}, \bibinfo{person}{Enyu Zhou}, {et~al\mbox{.}}} \bibinfo{year}{2023}\natexlab{}.
\newblock \showarticletitle{The rise and potential of large language model based agents: A survey}.
\newblock \bibinfo{journal}{\emph{arXiv preprint arXiv:2309.07864}} (\bibinfo{year}{2023}).
\newblock


\bibitem[Xian et~al\mbox{.}(2019)]%
        {xian2019reinforcement}
\bibfield{author}{\bibinfo{person}{Yikun Xian}, \bibinfo{person}{Zuohui Fu}, \bibinfo{person}{Shan Muthukrishnan}, \bibinfo{person}{Gerard De~Melo}, {and} \bibinfo{person}{Yongfeng Zhang}.} \bibinfo{year}{2019}\natexlab{}.
\newblock \showarticletitle{Reinforcement knowledge graph reasoning for explainable recommendation}. In \bibinfo{booktitle}{\emph{Proceedings of the 42nd international ACM SIGIR conference on research and development in information retrieval}}. \bibinfo{pages}{285--294}.
\newblock


\bibitem[Xu et~al\mbox{.}(2024)]%
        {xu2024generate}
\bibfield{author}{\bibinfo{person}{Yao Xu}, \bibinfo{person}{Shizhu He}, \bibinfo{person}{Jiabei Chen}, \bibinfo{person}{Zihao Wang}, \bibinfo{person}{Yangqiu Song}, \bibinfo{person}{Hanghang Tong}, \bibinfo{person}{Kang Liu}, {and} \bibinfo{person}{Jun Zhao}.} \bibinfo{year}{2024}\natexlab{}.
\newblock \showarticletitle{Generate-on-Graph: Treat LLM as both Agent and KG in Incomplete Knowledge Graph Question Answering}.
\newblock \bibinfo{journal}{\emph{arXiv preprint arXiv:2404.14741}} (\bibinfo{year}{2024}).
\newblock


\bibitem[Yao et~al\mbox{.}(2022)]%
        {yao2022react}
\bibfield{author}{\bibinfo{person}{Shunyu Yao}, \bibinfo{person}{Jeffrey Zhao}, \bibinfo{person}{Dian Yu}, \bibinfo{person}{Nan Du}, \bibinfo{person}{Izhak Shafran}, \bibinfo{person}{Karthik Narasimhan}, {and} \bibinfo{person}{Yuan Cao}.} \bibinfo{year}{2022}\natexlab{}.
\newblock \showarticletitle{React: Synergizing reasoning and acting in language models}.
\newblock \bibinfo{journal}{\emph{arXiv preprint arXiv:2210.03629}} (\bibinfo{year}{2022}).
\newblock


\bibitem[Zhang et~al\mbox{.}(2024c)]%
        {zhang2024generative}
\bibfield{author}{\bibinfo{person}{An Zhang}, \bibinfo{person}{Yuxin Chen}, \bibinfo{person}{Leheng Sheng}, \bibinfo{person}{Xiang Wang}, {and} \bibinfo{person}{Tat-Seng Chua}.} \bibinfo{year}{2024}\natexlab{c}.
\newblock \showarticletitle{On generative agents in recommendation}. In \bibinfo{booktitle}{\emph{Proceedings of the 47th International ACM SIGIR Conference on Research and Development in Information Retrieval}}. \bibinfo{pages}{1807--1817}.
\newblock


\bibitem[Zhang et~al\mbox{.}(2024d)]%
        {zhang2024agentcf}
\bibfield{author}{\bibinfo{person}{Junjie Zhang}, \bibinfo{person}{Yupeng Hou}, \bibinfo{person}{Ruobing Xie}, \bibinfo{person}{Wenqi Sun}, \bibinfo{person}{Julian McAuley}, \bibinfo{person}{Wayne~Xin Zhao}, \bibinfo{person}{Leyu Lin}, {and} \bibinfo{person}{Ji-Rong Wen}.} \bibinfo{year}{2024}\natexlab{d}.
\newblock \showarticletitle{Agentcf: Collaborative learning with autonomous language agents for recommender systems}. In \bibinfo{booktitle}{\emph{Proceedings of the ACM on Web Conference 2024}}. \bibinfo{pages}{3679--3689}.
\newblock


\bibitem[Zhang et~al\mbox{.}(2024b)]%
        {zhang2024lorec}
\bibfield{author}{\bibinfo{person}{Kaike Zhang}, \bibinfo{person}{Qi Cao}, \bibinfo{person}{Yunfan Wu}, \bibinfo{person}{Fei Sun}, \bibinfo{person}{Huawei Shen}, {and} \bibinfo{person}{Xueqi Cheng}.} \bibinfo{year}{2024}\natexlab{b}.
\newblock \showarticletitle{Lorec: Large language model for robust sequential recommendation against poisoning attacks}.
\newblock \bibinfo{journal}{\emph{arXiv preprint arXiv:2401.17723}} (\bibinfo{year}{2024}).
\newblock


\bibitem[Zhang et~al\mbox{.}(2024a)]%
        {zhang2024survey}
\bibfield{author}{\bibinfo{person}{Zeyu Zhang}, \bibinfo{person}{Xiaohe Bo}, \bibinfo{person}{Chen Ma}, \bibinfo{person}{Rui Li}, \bibinfo{person}{Xu Chen}, \bibinfo{person}{Quanyu Dai}, \bibinfo{person}{Jieming Zhu}, \bibinfo{person}{Zhenhua Dong}, {and} \bibinfo{person}{Ji-Rong Wen}.} \bibinfo{year}{2024}\natexlab{a}.
\newblock \showarticletitle{A survey on the memory mechanism of large language model based agents}.
\newblock \bibinfo{journal}{\emph{arXiv preprint arXiv:2404.13501}} (\bibinfo{year}{2024}).
\newblock


\end{thebibliography}
